%% file: main.tex
\documentclass[a4paper]{llncs}

\usepackage[T1]{fontenc}
\usepackage[english]{babel}
\usepackage{newtxtext}
\usepackage[varvw]{newtxmath}
\usepackage{graphicx}
\usepackage{booktabs}
\usepackage{multirow}
\usepackage{microtype}
\usepackage{xcolor}
\usepackage{url}
\usepackage[hidelinks]{hyperref}
\graphicspath{{figures/}}
\input{tables/generated_macros}

\newcommand{\repeatthanks}{\textsuperscript{$\star$}}

\newcommand{\dataset}{\textsc{StateBind-200}}
\newcommand{\numeric}{\textsc{Numeric}}
\newcommand{\calibrated}{\textsc{Calibrated}}
\newcommand{\physicalized}{\textsc{Physicalized}}
\newcommand{\factorial}{\textsc{Factorial StateBind}}

\newcommand{\R}{\mathcal{R}}
\newcommand{\C}{\mathcal{C}}
\newcommand{\Hset}{\mathcal{H}}
\newcommand{\vect}[1]{\boldsymbol{#1}}

\begin{document}

\title{StateBind: Exposing the Symbol-to-State Binding Gap in Visual Measurement}
\titlerunning{StateBind: Symbol-to-State Binding in Visual Measurement}
\author{Qing Mu\inst{1}\thanks{Equal contribution. Corresponding author: Qing Mu.}\orcidID{0009-0004-9571-2728} \and
Jiani Zheng\inst{2}\repeatthanks\orcidID{0009-0006-2835-5147}}
\authorrunning{Q. Mu and J. Zheng}
\tocauthor{Qing Mu and Jiani Zheng}
\institute{The Chinese University of Hong Kong, Shenzhen, Shenzhen, China\\
\email{qingmu@link.cuhk.edu.cn} \and
Tencent, Shenzhen, China\\
\email{ninajnzheng@tencent.com}}
\maketitle
\pagestyle{empty}
\thispagestyle{empty}

\begin{abstract}
Reliable vision systems should connect numerical claims to visible evidence. In visual measurement, this requires mapping candidate values to image-specific physical states. \dataset{} augments \DatasetImages{} real images with human-refined local geometry and matched \numeric{}, \calibrated{}, and \physicalized{} interfaces. Across a \PairedImages{}-image held-out evaluation and \FormalCalls{} controlled calls, physicalizing identical candidates raises three-model majority accuracy from \MajorityNumeric\% to \MajorityPhysicalized\%, a source-image-balanced gain of \MajorityBalancedGain{} points. Calibration and additional reasoning leave most of the gap intact; target localization can succeed despite incorrect value--lattice binding, with a larger burden for two-endpoint ruler spans. An independent \GeometryAuditImages{}-image geometry audit yields a \GeometryAuditTargetMedian{}-tick median target discrepancy and preserves the main effect under disagreement sensitivity.

We further propose \factorial{}, a training-free selective verifier. Two VLMs propose target and lattice components, an independent VLM validates their evidence, and valid combinations are resolved deterministically. A precision mode selected on 69 development images combines component validity, four-way agreement, and state separation. Applied unchanged to held-out evaluation, this high-precision evidence gate reaches \FactorialTestPrecision\% candidate precision across \FactorialTestImages{}/128 images (\FactorialImageCoverage\%) and \FactorialTestCoverage\% of ordered cells. The results identify instance-specific symbol-to-state binding as a distinct reliability bottleneck and provide an interpretable evidence gate for numerical visual claims.

\keywords{Vision-language models \and Visual measurement \and Visual grounding \and Multimodal reliability \and Selective verification}
\end{abstract}

\input{sections/01_introduction}
\input{sections/02_related_work}
\input{sections/03_task_and_data}
\input{sections/04_binding_gap}
\input{sections/05_method}
\input{sections/06_experiments}
\input{sections/07_discussion}
\input{sections/08_conclusion}

\clearpage
\bibliographystyle{splncs04}
\bibliography{references}

\end{document}

%% file: tables/generated_macros.tex
\newcommand{\DatasetImages}{200}
\newcommand{\DatasetFamilies}{11}

\newcommand{\HeldoutImages}{131}
\newcommand{\PairedImages}{128}
\newcommand{\ImageDistancePairs}{372}
\newcommand{\CallsPerInterface}{744}

\newcommand{\FormalCalls}{6,696}
\newcommand{\MajorityNumeric}{75.54}

\newcommand{\MajorityPhysicalized}{91.13}
\newcommand{\MajorityBalancedGain}{+15.30}
\newcommand{\MajorityGainLow}{10.35}
\newcommand{\MajorityGainHigh}{20.31}
\newcommand{\AllCorrectNumeric}{45.83}
\newcommand{\AllCorrectPhysicalized}{66.80}

\newcommand{\FactorialTestPrecision}{91.67}
\newcommand{\FactorialTestCoverage}{9.68}
\newcommand{\FactorialTestRouted}{72}

\newcommand{\FactorialTestImages}{36}
\newcommand{\FactorialImageCoverage}{28.12}

\newcommand{\GeometryAuditImages}{40}
\newcommand{\GeometryAuditFamilies}{11}
\newcommand{\GeometryAuditTargetMedian}{0.08}
\newcommand{\GeometryAuditWithinHalf}{37}
\newcommand{\GeometryAuditWithinOne}{38}
\newcommand{\GeometryAuditAxisMedian}{0.17}
\newcommand{\GeometryAuditTickSpacingMedianDelta}{4.6}
\newcommand{\GeometryAuditSensitivityGain}{+15.54}
\newcommand{\FourWayFamilyBootstrapSelection}{92.03}
\newcommand{\FrozenRuleBootstrapFeasible}{92.46}

%% file: sections/01_introduction.tex
\section{Introduction}
\label{sec:intro}

A numerical answer is trustworthy only when it can be traced to evidence in the current image. Visual measurement makes this requirement explicit: a model must identify the scale and boundary, reconstruct the local ticks, and connect each candidate to the state it denotes. Recent benchmarks show that vision--language models (VLMs) remain unreliable even when they recognize the instrument and unit~\cite{lin2026measurebench,jin2026scalebench}, but aggregate accuracy mixes perception, scale, localization, and grounding failures.

\begin{figure}[t]
  \centering
  \includegraphics[width=\textwidth]{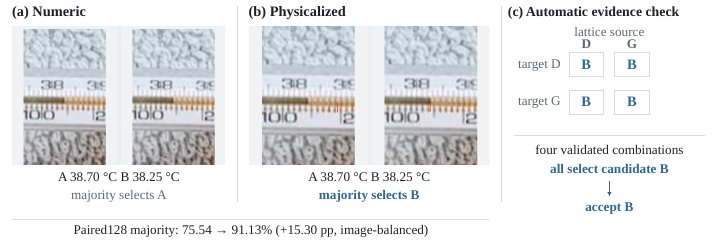}
  \Description{Three views of one thermometer example compare the same numerical candidates, candidate states drawn on the photographed scale, and an automatic check using four target-lattice combinations. The Numeric majority is incorrect and the Physicalized majority is correct.}
  \caption{\textbf{Same numbers, different access to their physical states.} Numeric presents two symbolic candidates; Physicalized renders the same values on the photographed scale. In this held-out example, the three-model majority changes from an incorrect to a correct choice. The automatic method tests the candidate across four validated target--lattice combinations. Source image adapted from MeasureBench~\cite{lin2026measurebench}, licensed under CC BY-SA 4.0; cropped and annotated by the authors.}
  \label{fig:teaser}
\end{figure}

We focus on the correspondence between a numerical hypothesis and its image-specific state. For candidates 38.70 and 38.25\,$^{\circ}$C, a conventional interface requires the model to construct each position on the photographed scale. A \emph{physicalized} interface draws those states instead. Image, question, values, order, and scoring remain matched; only the denoted states become visible. In Fig.~\ref{fig:teaser}, the three-model majority rejects the correct symbolic candidate but selects it after physicalization.

This discontinuity isolates a \emph{symbol-to-state binding gap}. The map from value $v$ to visible state $g_I(v)$ changes with each image's branch, direction, phase, spacing, perspective, and operator. \dataset{} isolates it on \DatasetImages{} real images with human-refined local geometry. Across three matched interfaces, two candidate orders, and three VLMs, balanced gains are 8.40, 16.28, and 16.08 points for Doubao, Gemini, and Qwen. Three-model majority gains \MajorityBalancedGain{} points, removing 63.7\% of its remaining Numeric errors at call level. Calibration stays near Numeric, while the effect persists across reasoning budgets and major instrument families.

Oracle Physicalized uses human geometry; automatic verification must estimate both target and value--scale map. A proposal may remain self-consistent when target and lattice share a displacement. We therefore introduce \factorial{}: Doubao and Gemini propose target and lattice components, Qwen validates their evidence, and all four source combinations are resolved without another holistic answer. Precision mode requires executable components, four-way candidate agreement, and four-tick separation. The rule is selected on Dev69 and applied unchanged to held-out evaluation.

Our contributions are threefold:
\begin{itemize}
  \item We augment 200 real MeasureBench images with human-refined local geometry and matched Numeric, Calibrated, and Physicalized interfaces, forming \dataset{}. Its main matrix contains \FormalCalls{} controlled calls over \ImageDistancePairs{} image--distance pairs, complemented by an independent geometry audit spanning all \GeometryAuditFamilies{} families.
  \item We identify a cross-model binding gap: physicalizing identical numerical candidates raises majority accuracy by \MajorityBalancedGain{} points. The gap persists after scale calibration and increased reasoning, separates from target localization, and is larger on two-endpoint ruler spans.
  \item We propose \factorial{}, a training-free selective verifier that validates independently proposed components, recombines their sources, and accepts decisions stable across valid combinations and sufficiently separated in the physical state space. Its held-out precision is \FactorialTestPrecision\% across \FactorialTestImages{} images and \FactorialTestCoverage\% of ordered cells.
\end{itemize}

%% file: sections/02_related_work.tex
\section{Related Work}
\label{sec:related}

\paragraph{Visual measurement.}
MeasureBench establishes real and synthetic instrument reading as a broad VLM challenge, while ScaleBench extends recognition across quantities and question forms~\cite{lin2026measurebench,jin2026scalebench}. BLINK likewise finds that elementary perception can resist language mediation~\cite{fu2024blink}. Measurement systems decompose geometry, scale, and readout~\cite{reitsma2024underpressure,pan2026rulernet}; TriSCA uses state-aware training to make dial readings consistent across appearance changes~\cite{hu2026trisca}. Our benchmark instead diagnoses general-purpose VLMs across heterogeneous one-dimensional instruments through matched symbolic and physical interfaces and an inference-time selective verifier.

\paragraph{Spatial numeracy and grounding.}
Region-aware models and datasets connect language to image locations at different granularities~\cite{kamath2021mdetr,you2024ferret,deitke2024molmo}. Set-of-Mark prompting externalizes visual references for general VLMs~\cite{yang2023setofmark}, although marker-based evaluation can be sensitive to rendering choices~\cite{feng2025visualpromptfragility}. SpatialVLM, GroundingME, and SPACENUM probe metric reasoning, localization, and numerical--spatial correspondence~\cite{chen2024spatialvlm,li2026groundingme,zhang2026spacenum}; coordinate tokenization and bidirectional correspondence provide complementary grounding formulations~\cite{zhu2026hitoken,zhang2026concor}. Our setting requires an instance-specific map on a continuous lattice: the same value can occupy different pixels in every photograph, and the correct state depends on scale branch, phase, spacing, direction, and operator. We fix one renderer and randomize candidate order so that Physicalized serves as a matched representation intervention rather than a marker search.

\paragraph{Consistency and multi-model verification.}
Reasoning traces and repeated-answer voting add language-side computation~\cite{wei2022chain,wang2023selfconsistency}, yet high agreement can still accompany shared error~\cite{bhat2026consistentwrong,zhang2025consensusentropy}. Visual self-correction likewise depends strongly on the verifier and model~\cite{liao2025selfcorrect}. CALIBRA frames physiological rate estimation as calibrated, evidence-grounded multi-agent verification~\cite{sakib2026calibra}. Our verifier instead diagnoses an image-specific symbol-to-lattice map, recombines target and lattice sources, and resolves candidate identities deterministically.

\paragraph{Selective prediction.}
Selective classification trades coverage for lower risk~\cite{elyaniv2010selective,geifman2017selective}. In VQA, learned peer confidence and consistency under question rephrasing can identify unreliable answers~\cite{dancette2023selectivevqa,khan2024consistency}. Here, visible component validity, cross-source candidate agreement, and state separation define an interpretable accept-or-fallback decision. This connects candidate reliability to the physical support for a numerical claim, aligning with broader efforts toward verifiable visual reasoning~\cite{xu2025vpbench}.

%% file: sections/03_task_and_data.tex
\section{Problem Formulation and StateBind-200}
\label{sec:data}

\subsection{Instance-Specific Symbol-to-State Binding}

Let $I$ be a real instrument image, $q$ its measurement question, and $\C=\{c_1,c_2\}$ two numerical candidates expressed in a common unit. A measurement is governed by an image-specific state map
\begin{equation}
  g_I : v \mapsto s,
  \label{eq:map}
\end{equation}
where $v$ is a numerical value and $s$ is its physical state on the photographed scale. In the simplest case $s$ is a scalar coordinate along an axis; for a span it is a pair of endpoint states. Unlike a semantic class label, $g_I$ changes with every image. It includes the selected scale branch, unit, direction of increase, local tick spacing and phase, perspective, and readout operator.

We instantiate two alternatives and a one-dimensional local state so the matched intervention is exact. Equations~\eqref{eq:map}--\eqref{eq:candidate} admit any finite candidate set and structured state for which a residual is defined.

The visual target is $s_t(I,q)$. Given a state residual $\R$, candidate verification is
\begin{equation}
  c^\star = \arg\min_{c\in\C} \R\!\left(g_I(c),s_t\right).
  \label{eq:candidate}
\end{equation}
For absolute or boundary readout, $\R$ is distance to the target interval; for a span, it compares the candidate with the difference between two endpoints. We use \emph{absolute position}, \emph{boundary position} (e.g., liquid surfaces or columns), and \emph{span difference}. A model may therefore estimate $s_t$ well while estimating $g_I$ poorly, or vice versa.

\subsection{Human-Refined Local State Geometry}

\dataset{} augments real-world MeasureBench images~\cite{lin2026measurebench}. It contains \DatasetImages{} images with a visible one-dimensional physical scale, spanning \DatasetFamilies{} instrument families including thermometers, cylinders, rulers, sphygmomanometers, syringes, flow meters, spirometers, rain and fuel gauges, a vernier caliper, and a weighing scale.

Each annotation records an axis; target interval or span endpoints; two adjacent minor ticks near the readout; tick value, unit, branch, visibility, and operator. Clicks are projected into the one-dimensional coordinate, with local ticks accommodating perspective. Across 53 auxiliary ticks not used to fit the local map from 24 multi-tick images, the two-tick affine map has 0.065-tick median error (mean 0.122; maximum 0.487). The 131-image extension was proposal-seeded to reduce clicking, but every record was modified and human-confirmed while model predictions and evaluator answers were hidden.

A second reviewer then redrew the geometry of a stratified \GeometryAuditImages{}-image audit spanning all \GeometryAuditFamilies{} families. Primary point coordinates remained hidden; tick semantics were prefilled and editable so the audit focused on spatial reproduction. All \GeometryAuditImages{} images remained included. Median target discrepancy was \GeometryAuditTargetMedian{} tick, with \GeometryAuditWithinHalf{}/\GeometryAuditImages{} cases within 0.5 tick and \GeometryAuditWithinOne{}/\GeometryAuditImages{} within one tick; median local-scale orientation difference was \GeometryAuditAxisMedian$^{\circ}$ and median tick-spacing difference was \GeometryAuditTickSpacingMedianDelta\%. Excluding the two cases beyond one tick leaves the majority effect at \GeometryAuditSensitivityGain{} points, compared with \MajorityBalancedGain{} on the complete evaluation.

\begin{table}[t]
  \centering
  \caption{\dataset{} paired-test protocol. Every planned cell contributes one successfully parsed final response; retries preserve its cell ID, and no output is filtered by predicted choice.}
  \label{tab:dataset}
  \small
  \setlength{\tabcolsep}{8pt}
  \input{tables/dataset_protocol.tex}
\end{table}

The first 69 images form \textsc{Dev69}; the later \HeldoutImages{} form the held-out pool (Table~\ref{tab:dataset}). After cell-level visibility and rendering checks, \PairedImages{} unique images contribute \ImageDistancePairs{} valid image--distance pairs: 127 at one tick, 125 at two ticks, and 120 at four ticks. Each pair is evaluated in both candidate orders, yielding \CallsPerInterface{} ordered cells per interface and model. Reliability configurations are selected on Dev69 and applied unchanged to Paired128; held-out labels are used only for final scoring.

\subsection{Matched Candidate Interfaces}
\label{sec:interfaces}

For each renderable image--distance pair with $d\in\{1,2,4\}$, a correct candidate is paired with an alternative displaced by $d$ local ticks. Primary and swapped orders are reused unchanged across interfaces:

\paragraph{Numeric.}
Image, question, and numerical candidates; the VLM infers both $s_t$ and $g_I(c)$.

\paragraph{Calibrated.}
Numeric plus the correct branch, unit, direction, and minor-tick value; no candidate location is drawn.

\paragraph{Oracle Physicalized.}
Candidates are rendered in matched panels at $g_I(c_1)$ and $g_I(c_2)$. Numerical identities, order, crop, and question remain matched. This diagnostic interface isolates candidate-state access; Sec.~\ref{sec:method} develops the automatic counterpart.

Models return structured A/B choices, scored by candidate identity after undoing order randomization. The matrix contains \CallsPerInterface{} calls per interface per model and \FormalCalls{} total calls.

\subsection{Evaluation Unit}

Raw accuracy uses all \CallsPerInterface{} ordered cells per interface and model. Treatment effects first average paired differences within each source image, then across images. Confidence intervals resample source images (20,000 samples per model; 50,000 for majority and family analyses). Tables distinguish call accuracy from image-balanced effects.

The companion repository\footnote{\url{https://github.com/mqingcs/StateBind}}
provides 200 geometry records, exact manifests for all 2,232 requests,
the 6,696-response ledger, proposal and validation records,
and deterministic rebuild scripts for all reported results.

%% file: tables/dataset_protocol.tex
\begin{tabular}{lr}
\toprule
Real images / instrument families & 200 / 11 \\
Development / held-out images & 69 / 131 \\
Unique paired-evaluation images & 128 \\
Valid image--distance pairs & 372 \\
Ordered cells per interface / model & 744 \\
Candidate offsets / orders & 1, 2, 4 ticks / 2 \\
Calls per model / main matrix & 2,232 / 6,696 \\
\bottomrule
\end{tabular}

%% file: sections/04_binding_gap.tex
\section{The Symbol-to-State Binding Gap}
\label{sec:gap}

\begin{table}[t]
  \centering
  \caption{\textbf{Matched interface comparison on Paired128.} Interface columns are call accuracies (\%). The final column is the source-image-balanced Physicalized-minus-Numeric effect in percentage points, so it need not equal the difference of rounded call accuracies.}
  \label{tab:main}
  \small
  \setlength{\tabcolsep}{4pt}
  \input{tables/main_results.tex}
\end{table}

\subsection{Physicalized Candidates Improve Verification}

Table~\ref{tab:main} reports the complete paired result. Doubao, Gemini, and Qwen improve by source-image-balanced effects of $+8.40$, $+16.28$, and $+16.08$ points. Gains remain positive at every candidate distance. The three-model majority rises from \MajorityNumeric\% to \MajorityPhysicalized\%; its balanced effect is \MajorityBalancedGain{} points, 95\% CI $[\MajorityGainLow,\MajorityGainHigh]$. At the call level, Physicalized removes 63.7\% of the majority errors remaining under Numeric.

\begin{figure}[t]
  \centering
  \includegraphics[width=\textwidth]{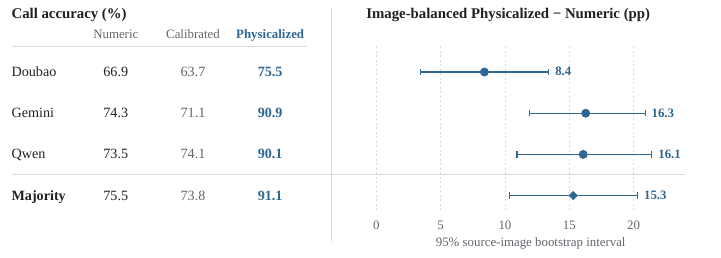}
  \Description{Two panels compare Numeric, Calibrated, and Physicalized call accuracy for Doubao, Gemini, Qwen, and the three-model majority, then show positive image-balanced Physicalized-minus-Numeric effects with bootstrap confidence intervals. Calibration stays near Numeric.}
  \caption{\textbf{Matched interface comparison.} Left: call accuracy for the three interfaces. Right: source-image-balanced Physicalized-minus-Numeric effects with 95\% source-image bootstrap intervals. Calibration remains near Numeric, while physicalizing candidate states produces a consistent gain.}
  \label{fig:ladder}
\end{figure}

All-model correctness rises from \AllCorrectNumeric\% to \AllCorrectPhysicalized\%, with 147 majority repairs and 31 breaks. Figure~\ref{fig:ladder} shows the same trajectory for each model. Calibrated remains close to Numeric: 63.71, 71.10, and 74.06\% versus 66.94, 74.33, and 73.52\%. Correct branch, unit, direction, and tick value therefore do not supply the missing absolute correspondence between the candidates and the photographed scale.

The result is not driven by one instrument type. Removing each major family leaves a $+12.06$ to $+18.19$-point majority effect, and every source-image bootstrap interval remains above zero. In particular, the gain is $+12.06$ points without rulers and $+18.19$ without thermometers, the largest family.

\subsection{Physicalization Gains Persist Across Reasoning Budgets}
\label{sec:reasoning}

Fresh45 provides matched effort matrices for additional model settings. Physicalized $-$ Numeric effects remain positive for Gemini at minimal/medium/high effort ($+16.30/\allowbreak +19.63/\allowbreak +18.15$ points), GPT-5.6 Terra at low/medium/high ($+16.30/\allowbreak +17.41/\allowbreak +14.81$), and Qwen at low/medium/extra-high ($+24.81/\allowbreak +25.56/\allowbreak +27.04$). GPT-5.6 Sol at medium effort independently rises from 81.85\% to 91.85\%. Low-effort Physicalized also exceeds high-effort Numeric by 10.00, 13.70, and 25.56 points for Gemini, Terra, and Qwen. Increasing language-side computation does not substitute for the missing image-specific state map.

\subsection{Complementary Probes of the Gap}

\begin{figure}[t]
  \centering
  \includegraphics[width=\textwidth]{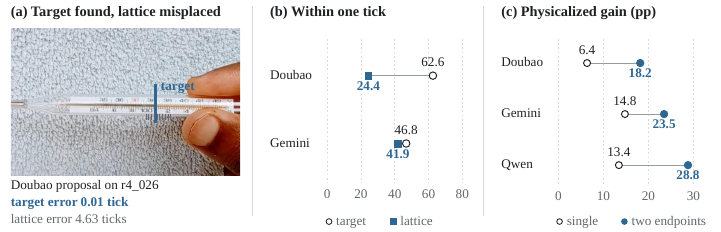}
  \Description{An annotated instrument image shows an accurately localized target beside a displaced scale lattice. Two charts compare target and lattice localization within one tick and show larger Physicalized gains for two-endpoint ruler spans than for single-state tasks.}
  \caption{\textbf{Complementary probes.} A real proposal can localize the measured boundary while placing the local lattice incorrectly. Aggregate target and lattice errors show the same separation, and Physicalized gains are larger for two-endpoint ruler spans than for single-state tasks. Source image: MeasureBench~\cite{lin2026measurebench}, CC BY-SA 4.0; cropped and annotated by the authors.}
  \label{fig:mechanism}
\end{figure}

The automatic proposers expose target and lattice estimates before human geometry is used for evaluation. On held-out images, Doubao localizes the target within one tick on 62.60\% of 123 images but reconstructs the lattice within one tick on only 24.39\% (median errors 0.45 versus 3.40 ticks). Gemini reaches 46.77\% and 41.94\% over 124 images. Thus, a target can be found even when the numerical anchors are bound to the wrong local lattice. Figure~\ref{fig:mechanism} shows both this component asymmetry and the larger effect on two-endpoint measurements.

\subsection{Two-Endpoint Ruler Spans Increase the Burden}

On 22 ruler-span images, Numeric accuracy ranges from 49.22\% to 57.81\% across Doubao, Gemini, and Qwen. Physicalization gains $+18.18$, $+23.48$, and $+28.79$ points, compared with $+6.37$, $+14.78$, and $+13.44$ on 106 single-state images. Explicitly instructing Doubao and Qwen to read both endpoints and subtract still leaves Physicalized ahead by $+19.70$ and $+35.61$ points. The result points to an upstream grounding bottleneck: two image states must first be aligned to the same scale.

%% file: tables/main_results.tex
\begin{tabular}{lrrrr}
\toprule
Model & Numeric & Calibrated & Physicalized & Balanced $\Delta_{P-N}$ \\
\midrule
Doubao Seed 2.1 Turbo & 66.94 & 63.71 & 75.54 & +8.40 \\
Gemini 3.6 Flash & 74.33 & 71.10 & 90.86 & +16.28 \\
Qwen3.8-Max & 73.52 & 74.06 & 90.05 & +16.08 \\
\midrule
Three-model majority & 75.54 & 73.79 & 91.13 & \textbf{+15.30} \\
\bottomrule
\end{tabular}

%% file: sections/05_method.tex
\section{Factorial StateBind}
\label{sec:method}

\begin{figure}[t]
  \centering
  \includegraphics[width=\textwidth]{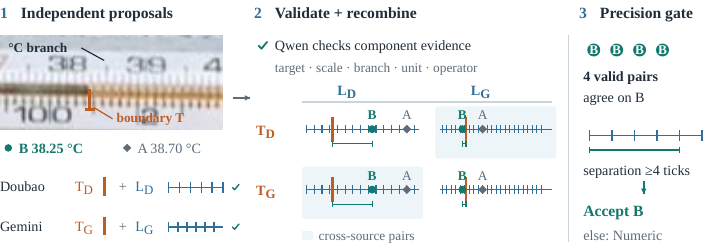}
  \Description{A three-stage diagram uses the existing thermometer example: Doubao and Gemini propose targets and scale lattices, Qwen validates component evidence, and a two-by-two matrix exchanges target and lattice sources. All four valid recombinations choose B, 38.25 degrees Celsius. The precision gate accepts only four-way agreement with at least four-tick candidate separation; otherwise the system retains Numeric.}
  \caption{\textbf{Factorial StateBind.} Qwen validates Doubao ($D$) and Gemini ($G$) target--lattice proposals. Projected schematic grids show four recombinations: shading marks cross-source pairs; brackets show target-to-B residuals. Precision mode requires four valid agreeing pairs and at least four-tick candidate separation. Source: MeasureBench~\cite{lin2026measurebench}, CC BY-SA 4.0; cropped and annotated by the authors.}
  \label{fig:architecture}
\end{figure}

Oracle Physicalized uses human geometry. An automatic verifier must instead estimate the target state and the value--scale mapping from the image, while recognizing when those estimates support a stable choice. The pipeline (Fig.~\ref{fig:architecture}) has three steps: factor two proposals into target and lattice components, verify each component once, and accept only choices invariant to exchanging component sources.

\subsection{Propose, Canonicalize, and Validate}

Two VLM proposers $D$ and $G$ receive only $(I,q,\C)$. We write response $r$ as $P_r=(T_r,L_r,o_r)$: target geometry, a local scale lattice (axis, ticks, anchors, tick value, unit, branch, and direction), and its readout operator. Coordinates use a normalized $[0,1000]^2$ frame and are rescaled to native pixels.

The proposer orients axis direction $\vect u$ toward increasing values. Projection onto $\vect u$ signs $\delta_p$ numerically. From projected ticks $p_k$, anchors $(p_a,v_a)$, and minor-tick value $\delta_v$, the local map is
\begin{equation}
 \delta_p=\operatorname{median}_k(p_{k+1}-p_k),\quad
 \phi=\operatorname{median}_a\!\left(p_a-\frac{v_a}{\delta_v}\delta_p\right),\quad
 \hat p(v)=\phi+\frac{v}{\delta_v}\delta_p .
 \label{eq:auto-map}
\end{equation}
Unit conversion is deterministic. Qwen acts as an independent component verifier: it inspects rendered target, anchor, lattice, branch, and operator evidence but does not output an A/B candidate. A proposal is executable when the required components are visually supported, tick spacing is positive, units are compatible, and the target and lattice use the same operator. Human geometry and evaluator answers are not inputs to this process.

\subsection{Cross-Source Recombination}

The two proposals yield four target--lattice combinations,
\begin{equation}
 \Hset=\{(T_D,L_D),(T_D,L_G),(T_G,L_D),(T_G,L_G)\}.
 \label{eq:factorial}
\end{equation}
Combination $(T_i,L_j)$ is executable only when both components pass validation and $o_i=o_j$. Let $\hat z_{ij}=\Psi_{o_i}(T_i,L_j)$ be the reading obtained by interpreting target $T_i$ in lattice $L_j$ under their shared operator. For a single-state task, $\Psi$ maps the target interval through $L_j$; for a span, it returns the absolute difference between the two mapped endpoints. Invalid combinations are omitted. Every valid combination deterministically selects
\begin{equation}
 \hat c_{ij}=\arg\min_{c\in\C}\R(c;\hat z_{ij}),\qquad
 \rho(c)=\frac{\sum_{(i,j)\in\Hset_{\rm valid}}\mathbf{1}[\hat c_{ij}=c]}
 {|\Hset_{\rm valid}|},
 \label{eq:choice}
\end{equation}
where $\R$ is distance from a candidate to a single-state interval or to the estimated endpoint difference. The method therefore changes which source supplies the target and lattice without asking a VLM to make four new holistic guesses.

\subsection{Precision Mode}

The general method exposes component validity, candidate support $\rho$, state separation, and dispersion. Let $\Delta_{\C}=|c_1-c_2|/\delta_v$ denote candidate separation in local ticks. For this benchmark we predefine nine selective rules on Dev69: three agreement requirements (4/4; unanimous over at least three valid combinations; three-of-four majority), crossed with minimum separations of 1, 2, and 4 ticks. Maximum coverage subject to at least 95\% development precision selects
\begin{equation}
 |\Hset_{\rm valid}|=4,\qquad \rho(\hat c)=1,\qquad
 \Delta_{\C}\ge4~\text{ticks}.
 \label{eq:precision}
\end{equation}
This precision mode routes 50 of 414 development ordered cells at 98.00\% precision and is applied unchanged to Paired128; cells not routed retain their Numeric reader. Across 20,000 source-image bootstrap samples, the four-way agreement family is selected in \FourWayFamilyBootstrapSelection\% of replicates, and the frozen four-tick rule satisfies the 95\% precision requirement in \FrozenRuleBootstrapFeasible\%. The winning separation varies among one, two, and four ticks, identifying four-source agreement as the stable rule family and four ticks as its high-precision operating point.

Self agreement uses $(T_D,L_D)$ and $(T_G,L_G)$ and can preserve a target--lattice displacement shared within each proposal. The off-diagonal pairs $(T_D,L_G)$ and $(T_G,L_D)$ break that coupling. Full recombination retains both views and asks whether the final candidate is stable across every component source.

%% file: sections/06_experiments.tex
\section{Automatic Selective Verification}
\label{sec:automatic}

\begin{table}[t]
  \centering
  \caption{\textbf{Same-support attribution.} All rows use four executable target--lattice combinations and four-tick candidate separation. Dev support contains 78 ordered cells; held-out support contains 116. Cell coverage is measured over all 744 held-out cells.}
  \label{tab:selective}
  \small
  \setlength{\tabcolsep}{4pt}
  \input{tables/selective_results.tex}
\end{table}

\subsection{Evaluation Protocol}

Doubao and Gemini independently propose image-level geometry for all 200 images. Existing Qwen records validate the target, anchors, lattice, branch, and operator of each proposal. These components are reused across candidate distances and orders; DD, DG, GD, and GG are deterministic recombinations and require no additional VLM call. Human geometry is joined only for evaluation. All reliability configurations are selected on Dev69 and applied unchanged to Paired128.

The main representation matrix uses Doubao with reasoning disabled and temperature 0, Gemini at medium thinking, and Qwen with thinking disabled. Candidate precision is correctness among routed cells; cell coverage is routed cells divided by all 744 ordered cells. Image coverage counts distinct routed images among Paired128.

\subsection{What Does Full Recombination Add?}

Table~\ref{tab:selective} compares rules on a common support: four executable combinations at four-tick candidate separation. A single development-selected constituent (GG) reaches only 77.59\% held-out precision. Requiring a three-of-four vote raises it to 82.61\%; self agreement reaches 81.82\%. Off-diagonal cross agreement is strongest on held-out evaluation at 92.11\%, showing that exchanging component sources carries a useful reliability signal. Full agreement is the only rule that reaches the 95\% development criterion, moving from 98.00\% on Dev69 to 91.67\% held-out precision. Cross and Full route different subsets.

Component validation and candidate separation are both necessary for the selected precision. At four ticks, full agreement without Qwen validation routes 136 held-out cells at 77.94\%; validation yields a 72-cell route at 91.67\%. Among validated full-agreement cells, precision rises with separation: 68.57\% at one tick, 75.68\% at two, and 91.67\% at four. This gives the selected route a direct evidence-and-margin interpretation; recombination itself remains independent of the chosen operating point.

\begin{figure}[t]
  \centering
  \includegraphics[width=0.90\textwidth]{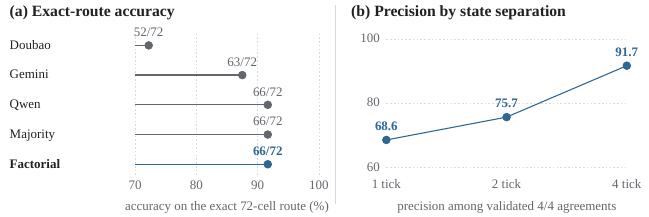}
  \Description{Two panels show accuracy on the fixed 72-cell route and candidate precision at different state separations. Numeric reader results are compared with the automatic verifier on identical routed cells; stronger state separation raises precision among four-way agreements.}
  \caption{\textbf{Exact-route behavior and state separation.} Left: the exact 72-cell route compared with its Numeric readers; labels report correct/routed counts. Right: candidate separation sharpens precision among validated four-way agreements.}
  \label{fig:frontier}
\end{figure}

\subsection{Exact-Route Readers and Fallback}

The precision mode reaches \FactorialTestImages{} of 128 images (\FactorialImageCoverage\%) across five instrument families, routing \FactorialTestRouted{} of 744 ordered cells (\FactorialTestCoverage\%). Figure~\ref{fig:frontier} compares the readers on exactly those cells: Doubao, Gemini, Qwen, and three-model majority are correct on 52, 63, 66, and 66 of 72, while the verifier is correct on 66. Qwen's component verifier never outputs the candidate; the validated structure instead transfers an evidence-supported choice to a weaker upstream reader.

Relative to Doubao Numeric on the routed cells, the verifier makes 18 repairs and 4 breaks, reducing routed error by 70.0\%. With Numeric fallback, overall call accuracy rises from 66.94\% to 68.82\% ($+1.88$ points). For Gemini it makes 7 repairs and 4 breaks, reduces routed error by 33.3\%, and adds $+0.40$ overall points. Qwen has 4 repairs and 4 breaks, so its net accuracy is unchanged. The same candidate-free evidence gate therefore improves weaker readers and preserves the strongest reader's full-set accuracy.

\subsection{System Cost and Reuse}

The pipeline uses at most four image-level VLM calls per image: two proposals plus completed Qwen validations for 191 Gemini and 184 Doubao frames, yielding 775 stored records. Components are computed once per image and reused across all distance/order cells. Recorded usage totals 1.01M input, 183k output, and 390k provider-reported reasoning tokens; recombination and routing are deterministic offline operations.

%% file: tables/selective_results.tex
\begin{tabular}{lrrrrr}
\toprule
Rule & Dev c/r & Dev prec. & Test c/r & Test prec. & Cell cov. \\
\midrule
Single constituent & 69/78 & 88.46 & 90/116 & 77.59 & 15.59 \\
3-of-4 vote & 53/59 & 89.83 & 76/92 & 82.61 & 12.37 \\
Self agreement & 53/57 & 92.98 & 72/88 & 81.82 & 11.83 \\
Cross agreement & 49/52 & 94.23 & 70/76 & 92.11 & 10.22 \\
\textbf{Full factorial} & 49/50 & 98.00 & 66/72 & 91.67 & 9.68 \\
\bottomrule
\end{tabular}

%% file: sections/07_discussion.tex
\section{Discussion}
\label{sec:discussion}

\paragraph{What does Physicalization diagnose?}
The matched interfaces place much of the error in constructing $g_I(v)$. Models often recognize the unit and locate the boundary, yet place a candidate on the wrong local lattice. Calibration supplies scale semantics rather than instance-specific phase and position; extra reasoning uses the same incomplete representation. Physicalized makes the missing correspondence visible.

\paragraph{Why does component recombination help?}
A proposer can shift target and lattice together while remaining internally coherent. Exchanging sources tests whether the inferred candidate survives either substitution, while validation ensures that each substitution is visually supported.

\paragraph{When is selective verification useful?}
Oracle Physicalized reveals representational headroom; the automatic verifier asks where image-derived geometry supports a transferable decision. The redraw audit strengthens the former, while validation and cross-source agreement define the latter. Precision mode reaches 28.1\% of images across five families, reduces routed error for Doubao and Gemini, and preserves Qwen's full-set accuracy. Fixing two candidates and one renderer keeps the intervention causal; the formulation covers point, boundary, and two-endpoint measurements across 11 families.

%% file: sections/08_conclusion.tex
\section{Conclusion}

\dataset{} isolates an instance-specific binding gap: physicalizing matched candidates yields a \MajorityBalancedGain{}-point majority gain beyond calibration and additional reasoning, supported by independent coordinate redraws. Recombining validated target and lattice evidence, the automatic verifier reaches \FactorialTestPrecision\% on \FactorialTestRouted{} cells from \FactorialTestImages{} held-out images. Together, these results make numerical visual claims traceable to physical state.

\begin{credits}
\subsubsection{\discintname}
The authors have no competing interests to declare that are relevant to the content of this article.
\end{credits}